\documentclass[letterpaper, 10 pt, conference]{ieeeconf}  

\IEEEoverridecommandlockouts                              

\usepackage{graphics} 
\usepackage{epsfig} 
\usepackage{mathptmx} 
\usepackage{times} 
\usepackage{amsmath} 
\usepackage{amssymb}  
\usepackage{cite}                      
\usepackage{tabu}                      
\usepackage{booktabs}                  
\usepackage{url}
\usepackage{array,multirow}
\usepackage{tabularx}
\usepackage{subcaption}
\usepackage{url}
\usepackage[colorlinks,allcolors=black]{hyperref}
\let\labelindent\relax

\usepackage{enumitem}

\title{\LARGE \bf

Reality Fusion: Robust Real-time Immersive Mobile Robot Teleoperation with Volumetric Visual Data Fusion}
\author{Ke Li$^{1, 2}$, Reinhard Bacher$^{1}$, Susanne Schmidt$^{2}$, Wim Leemans$^{1}$, Frank Steinicke$^{2}$ 
\thanks{$^{1}$ Ke Li, Reinhard Bacher, and Wim Leemans are with the accelerator division at the Deutsche Elektronen Synchrotron DESY, Notkestraße 85, 22607 Hamburg, Germany. Contact email: {\tt\small ke.li1@desy.de}}%
\thanks{$^{2}$ Ke Li, Susanne Schmidt, and Frank Steinicke are with the Human-Computer Interaction group at Hamburg University, Vogt-Kölln-Straße 30, 22527 Hamburg, Germany.}%
\thanks{This work was supported by DASHH (HELMHOLTZ Graduate School for the Structure of Matter) with the Grant-No. HIDSS-0002.}
}

\begin{document}

\maketitle
\thispagestyle{empty}
\pagestyle{empty}

\begin{abstract}

We introduce \emph{Reality Fusion}, a novel robot teleoperation system that localizes, streams, projects, and merges a typical onboard depth sensor with a photorealistic, high resolution, high framerate, and wide FoV rendering of the complex remote environment represented as 3D Gaussian splats (3DGS). Our framework enables robust egocentric and exocentric robot teleoperation in immersive VR, with the 3DGS effectively extending spatial information of a depth sensor with limited FoV and balancing the trade-off between data streaming costs and data visual quality. We evaluated our framework through a user study with 24 participants, which revealed that \emph{Reality Fusion} leads to significantly better user performance, situation awareness, and user preferences. To support further research and development, we provide an open-source implementation with an easy-to-replicate custom-made telepresence robot, a high-performance virtual reality 3DGS renderer, and an immersive robot control package. \footnote{\url{https://github.com/uhhhci/RealityFusion}}

\end{abstract}

 
\section{INTRODUCTION}


In recent years, the rapid development of extended reality (XR) technology has presented enormous opportunities for robot teleoperation and telepresence systems \cite{VAMHRI2020}. The possibility to display 3D spatial cues about the robot's environment to the operators through an immersive head-mounted display (HMD) enables a remote ``telepresence'' experience that has the potential to significantly improve the operator's task performance.  However, building a robust real-time immersive robot teleoperation system is still faced with many technical challenges. On one hand, real-time 3D data capturing and streaming to the HMD is crucial in providing operators with accurate situational awareness of the robot's environment. On the other hand, spatial data streaming and processing presents a trade-off between visual quality and processing latency. While telepresence systems using omnidirectional camera \cite{02_Redirected_Walking_360_Robot} or multi-camera setups \cite{multi_view_driving_example} can provide operators with a high level of immersion and detailed information of the robot's environment, the latency in streaming, processing, and rendering these 3D data introduce the undesirable effect of VR motion sickness \cite{02_Redirected_Walking_360_Robot} and a delay in robot control and intervention  \cite{04_MR_CERN_ROBOT_point_cloud}. Due to the limitation of robots' payload and the limited computational resources of an embedded system, many real-world mobile platforms can only provide operators with real-time visual feedback through low-cost sensor setups such as a single 2D video camera or a single stereo depth camera. These sensors often capture visual information with a restricted field of view (FoV), which hinders users from observing the remote environment from obscured angles and therefore restricts them from establishing a concrete mental model of the robot's surroundings.

\begin{figure}
    \centering
    \includegraphics[width=1\linewidth]{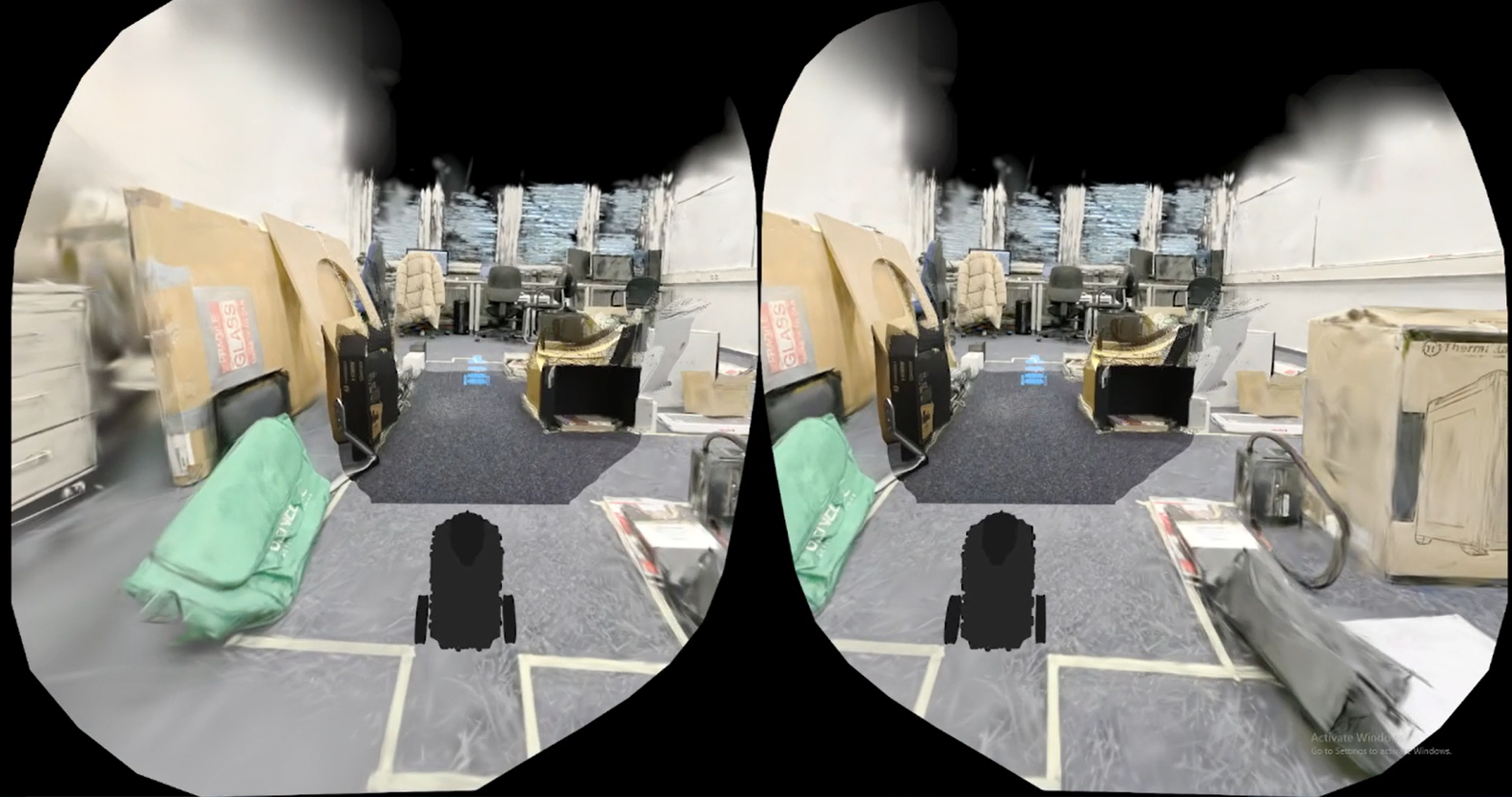}
    \caption{ Real-time point points from the robot's onboard RGBD sensor are localized, streamed, and projected onto a photorealistic scene of the remote environment represented as 3DGS in immersive VR.  }
    \label{fig:teaser}
\end{figure}

In this work, we propose \emph{Reality Fusion}, a novel immersive robot teleoperation framework that implements a multi-modal data fusion method to address the issue of spatial data streaming trade-offs. As Figure \ref{fig:teaser} illustrates, our framework first introduces a system that can render high resolution, high framerate, wide FoV, and photorealistic 3D scenes of complex environments that are represented as 3D Gaussian splats (3DGS)\cite{06_3DGS} in VR. Using this type of photorealistic scene representation as the digital replica of the robot's remote environment, the VR system enables offline visualization of complex scenarios that are typically difficult to accurately model with conventional 3D meshes or point clouds. 
To enable operators to view the real-time status of the remote environment, we introduce a data fusion method, where real-time point cloud from the onboard RGBD sensor is localized, streamed, projected, and merged with the 3DGS environment, with the 3DGS rendering of the remote environment effectively extending the FoV of the RGBD point cloud. Such a teleoperation system design allows us to implement a lightweight, flexible, and cost-efficient immersive telepresence mobile robot with only a single RGBD sensor and a small embedded system.

We systematically benchmarked our framework through a controlled user study experiment, in which 24 participants performed a mobile robot navigation task through a real-world maze. Furthermore, we compare the efficiency of egocentric and exocentric teleoperation, from which we discuss the advantages and trade-offs of these two teleoperation modes to derive further design insights for immersive telepresence systems using \emph{reality fusion}. 

    


    
 

\section{RELATED WORK}


\subsection{Immersive Robot Telepresence and Teleoperation}


According to Adamides et.al \cite{08_robot_teleop_usability_guideline}, one central aspect of an efficient teleoperation system is its capability to provide operators with a high level of situation awareness of the robot's surroundings. With the rapid advancement of VR technology, abundant research demonstrated that teleoperation through an immersive HMD can significantly enhance users'  situation awareness and task performance compared to traditional 2D display \cite{01_ego_exo_init_comparison, 02_Redirected_Walking_360_Robot, 04_merge_point_clouds, 04_MR_CERN_ROBOT_point_cloud}.  

A key challenge in building an immersive robot teleoperation system is to provide operators with high-quality and low-latency volumetric visual feedback of the robot's environment.  Ferland et al \cite{01_ego_exo_init_comparison}'s teleoperation system can perform a stereoscopic projection of a binocular camera into the user's world space, allowing for 6 Degrees of Freedom (DoF) exocentric immersive teleoperation in VR. However, operators' situation awareness can be largely restricted by the narrow FoV of the onboard camera. Although several previous works investigated using a multi-camera setup \cite{04_MR_CERN_ROBOT_point_cloud} or streaming videos from an omnidirectional camera \cite{02_Redirected_Walking_360_Robot}, these solutions introduce a significant increase in data streaming latency, leading to undesirable effect of VR motion sickness \cite{02_Redirected_Walking_360_Robot} and delay in robot control and intervention \cite{04_MR_CERN_ROBOT_point_cloud}. Another approach suggests displaying real-time mesh reconstruction results to the operator using dynamic SLAM algorithms \cite{VR-system-mesh}. However, 3D reconstruction with dynamic SLAM can introduce temporal delays of several seconds, making it unsuitable for applications requiring immediate intervention by the operator. Tefera et al. \cite{iros_gaze_MR_robotics} propose reducing live point cloud streaming bandwidth through a foveated point cloud segmentation and streaming framework, but this could introduce visual degradation and aliasing effects during third-person teleoperation, possibly reducing user flexibility and control options.
Various prior robot teleoperation systems use a cockpit-like design, integrating multiple sensor data sources—including 2D videos and 3D point clouds—into immersive user interfaces within a simulated control-room environment \cite{iros_humanoid_nasa_teleoperation, iros_VR_Control_Room_Mobile_Spot, iros_humanoid_robot_ego_exo}. These systems, similar to our approach, enable monitoring of a robot's movements from both egocentric and exocentric perspectives \cite{iros_humanoid_robot_ego_exo}. However, cockpit-like teleoperation frameworks typically use a large, immersive 3D space to display a wide range of visual information, which can vary in latency and spatial dimensions. This can lead to confusing interface designs and cognitive overload for users.
Our work extends previous teleoperation systems by combining limited online volumetric data with a high-quality offline photorealistic 3D representation of the environment. The offline representation serves as a contextual guideline to enhance operators' immersion, while the online data provides primary visual feedback for monitoring the robot's surroundings. This creates a more robust data streaming solution and a more coherent visualization style for real-time remote control. 

\subsection{3D Representations for Robot Teleoperation}

Another key challenge in immersive robot teleoperation is creating robust 3D representations to visualize the remote environment. Currently, most immersive robot teleoperation systems use conventional explicit representations such as point clouds and meshes to visualize the remote environment. However, 3D meshes for robot teleoperation are typically created with dynamic SLAM algorithms for real-time feedback \cite{VR-system-mesh}, which can yield inaccurate results with complex geometries, making them unsuitable to represent complex real-world conditions such as an industrial facility. Although point cloud is also a popular 3D representation in immersive robot teleoperation, such discrete representations introduce holes and occlusions, degrading the visual quality of the 3D representation and restricting users' understanding of the modeled environments \cite{04_merge_point_clouds}.  

The latest breakthrough in photorealistic scene rendering proposes a radiance field representation \cite{NeRF}. By storing such radiance fields in the form of 3D Gaussians to encode the directional optical property of a continuous volume space, radiance field rendering through 3D Gaussian splatting enables robust and accurate replication and real-time visualization of the appearance of our complex realities \cite{06_3DGS}. Despite the recency of its development, several works already investigated its applicability in robot localization and mapping \cite{Yan2023GSSLAMDV}. However, due to the early stage of research of 3DGS for robotics, existing 3DGS SLAM methods result in degraded visual quality or an increase in rendering frame timing, making them unsuitable for real-time immersive VR applications. Therefore, our framework focuses on developing an efficient integration of the current 3DGS method to an immersive robot teleoperation system for rendering 3DGS models offline rather than reconstructing dynamic SLAM 3DGS visual feedback online. Nonetheless, this is, to our best knowledge, the first usable robot teleoperation system based on immersive photorealistic rendering in VR.

\section{Reality Fusion}

We define the term ``\emph{Reality Fusion}" as the merging of two photorealistic 3D representations of a real-world environment with the purpose of data augmentation by combining the complementary features of different types of spatial data and naturally integrating them into a coherent spatial user interface (UI). In this work,  we develop a \emph{reality fusion} method that can combine real-time 3D projection of a stereo camera and the 3DGS of the real world to create a fully immersive telepresence experience for robot operators. This section presents the design goal and theoretical background related to such a \emph{reality fusion} method.  

\subsection{Design Goals}

Our framework specifically targets a typical application scenario of a robot teleoperation system for remote industrial facility inspection, where human access to the facility is limited due to various hazards or operation constraints, however, the environment of the facility is static and is unlikely to undergo immediate large structural change. A typical example of such an environment is a particle accelerator tunnel such as the large hadron collider (LHC) at the European Organization for Nuclear Research (CERN), as articulated in the work of Szczurek et al \cite{04_MR_CERN_ROBOT_point_cloud}. Building an effective and robust immersive teleoperation system for mobile robot navigation for such an application needs to consider the following design goals:

\begin{enumerate}[label=\textbf{G\arabic*},start=1,itemsep=0mm]
    \item The mobile robot needs to be flexible and lightweight, capable of visiting areas such as narrow gaps between facility components.
    \item The operator needs to have high situation awareness and spatial orientation of the robot's environment for task planning and navigation control.  
    \item The operator needs to receive real-time visual feedback on the robot's current status with minimum latency to perform timely intervention and robot control. 
    \item The implementation of the VR application needs to enable intuitive control of the robot without introducing undesirable effects such as motion sickness. 
\end{enumerate}


\subsection{3D Gaussian Splatting}

To fulfill the requirement of \textbf{G2}, our framework adapts 3DGS models for photorealistic scene rendering in immersive VR \cite{06_3DGS}. 
According to Kerbl et al, given an initial set of sparse points and camera poses estimated from a set of 2D images of a scene, the opacity and color of a real-world 3D volume can be represented as a set of 3D Gaussians optimized through Gaussian density control and gradient descent \cite{06_3DGS}. Each 3D Gaussian represents a part of the 3D volume with its position, rotation, scale, covariance matrix, as well as spherical harmonic coefficients which encode the view-dependent radiance field values for photorealistic rendering. The point blending method of 3D Gaussians (a.k.a splatting) can overlay a pixel by blending a summation of radiance contribution from each point to achieve state-of-the-art photorealistic view synthesis and rendering performance. As a result, a VR 3DGS system can rapidly duplicate and visualize the appearance of complex realities, enabling robot operators to be fully immersed in the remote environment. 

\subsection{Stereoscopic 3D Projection in World Space}

Rendering a 3DGS model can only provide offline passive visual feedback of the environment. To fulfill the requirement of \textbf{G3}, we integrate stereoscopic 3D projection of a depth camera to provide operators with real-time spatial feedback. 
Given a stereo camera, depth information ($d$) can be estimated based on the correspondence established from the two camera views. Given a homogeneous pixel vector with depth estimation $P_d=[x_d, y_d, d, 1]$, it's 3D coordinate $P_C=[x_c,y_c,z_c,w]$ can be calculated using stereoscopic 3D projection transformation \cite{01_ego_exo_init_comparison}: 

\begin{equation}
\begin{bmatrix}
    x_c        \\
    y_c        \\
    z_c        \\
    w        \\
\end{bmatrix} = 
\begin{bmatrix}
    a   & 0 & 0 & 0 \\
    0   & 1 & 0 & 0 \\
    0   & 0 & 0 & f \\
    0   & 0 & - \frac{1}{b} &0 \\
\end{bmatrix}
\begin{bmatrix}
    x_d        \\
    y_d        \\
    d        \\
    1        \\
\end{bmatrix}
\end{equation}

Here, $b$ is the stereo camera baseline, $a$ is the camera aspect ratio, and $f$ is the camera focal length, all of which could be obtained through camera calibration. Then, $P_C$ can be converted to homogeneous 3D coordinate $P_h$ through perspective division. 

To calculate the point cloud's 3D coordinate in world space $P_W =[x,y,z,1]$,  we transform $P_h$ through a transformation matrix obtained using the view projection matrix of the virtual camera $M_{VP}$ and transformation matrix of the stereo camera $M_{camera}$ which describes the tracked camera's position and orientation in global world space:

\begin{equation}
\label{eq:global_transform}
    P_W = M_{VP} \times T_{offset} \times M_{camera} \times P_h .
\end{equation}

As the robot's tracked position has a different center of origin than the actual position of the camera, an additional translation matrix $T_{offset}$ obtained through manual calibration is applied to compensate for the translation offset. 

As \emph{reality fusion} only depends on a single depth camera for receiving real-time visual feedback, it reduces the overall required payload on the robot (\textbf{G1}). In addition, the fusion of both a 3DGS model and real-time point clouds creates a coherent and natural visual appearance with low streaming and rendering latency  (\textbf{G4}). Moreover, both types of volumetric data enable 6DoF changes of perspective and robust exocentric robot control, making it easy for users to adjust their viewpoints and increasing the flexibility of robot planning and navigation control tasks (\textbf{G4}).  



\section{FRAMEWORK IMPLEMENTATION} \label{sec:system implementation}

\begin{figure*}
    \centering
    \includegraphics[width=1\linewidth]{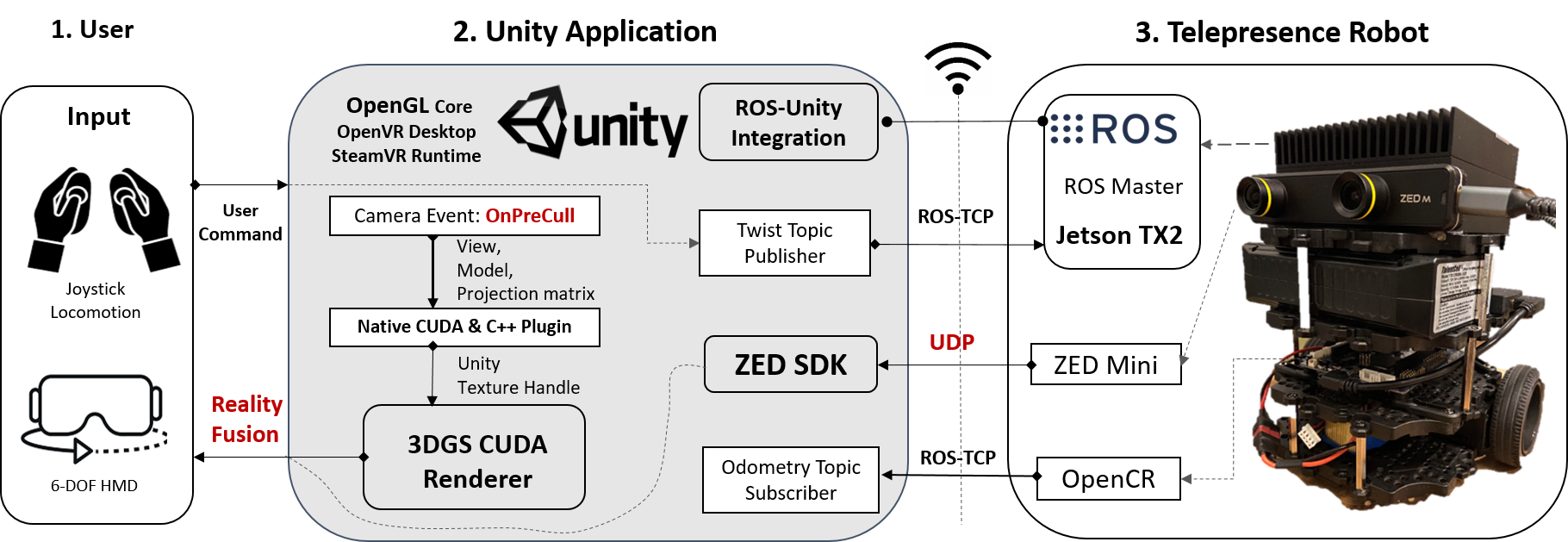}
    \caption{System overview of our immersive teleoperation framework which includes: 1. the operator equipped with a VR-HMD and sends command to the remote robot via VR controller inputs; 2. a Unity application which manages robot control logic, communicates with remote ROS endpoint, and perform data fusion and graphics rendering; 3. a custom-built telepresence mobile robot equipped with a SBC and a RGBD sensor. }
    \label{fig:system_overview}
\end{figure*}

As shown in Figure \ref{fig:system_overview}, our framework consists of three main components: (i) the robot operator equipped with a VR HMD and VR controllers for interaction with the spatial UI; (ii) a Unity application that handles the robot control logic, communicates with remote ROS endpoint, and perform high-performance graphics rendering through native CUDA and C++ plugins; and (iii) a telepresence robot with an onboard visual sensor providing real-time feedback to the remote operator. This section presents relevant details of our framework and implementations. 

\subsection{The Telepresence Robot}

\textbf{Overview}  We designed and developed a compact, lightweight, and modulized telepresence mobile robot that can be easily reassembled and replicated from commercially available hardwares. As shown in Figure \ref{fig:system_overview}, the robot is modified from the open-source Turtlebot 3 burger robot platform. Our custom-made robot has three core hardware components: an OpenCR board for low-level control of the robot's motion, a single ZED Mini stereo camera as the spatial vision sensor, and a single-board computer (SBC) as the robot's central computing unit. As with the original Turtlebot burger, the robot has two differential wheel drives with a maximum linear speed of $ 0.22 m / s$ and a maximum angular speed of $2.84 rad / s$.

\textbf{The Single-board Computer}\quad
The single-board computer (SBC) is an edge AI device based on an Nvidia TX2 NX, a high-performance embedded system with an accelerated 256-core NVIDIA Pascal GPU with CUDA version 10.2. The ZED Box runs on Ubuntu 18.04.6 LTS, Jetpack 4.6, ZED SDK version 3.7.3, and ROS Melodic. The SBC is powered by a TalentCell 72W 100WH power bank. 

\textbf{High-Resolution Stereo Camera and Video Streaming }\quad
A ZED Mini stereoscopic camera is mounted facing the forward direction of the robot. Stereoscopic videos are streamed at HD720 resolution with a vertical FoV of 54° and a horizontal FoV of 85°. The video stream is wirelessly sent via a local 5G network to the VR device through a user datagram protocol (UDP) with the ZED SDK, where streaming latency is minimized through highly optimized GPU video encoding and decoding processes. From a similar video streaming configuration, we can estimate the motion-to-photon latency of such a setup to be $153.47 \pm 33.33 ms$ at 30 frames per second (fps) \cite{MRTunneling} within a local 5G network. In addition, the ZED Mini camera also has a built-in 6DoF inertial measurement unit (IMU) to obtain $M_{camera}$ for accurate motion tracking. 

\subsection{Unity 3DGS VR Renderer}


\textbf{Overview}\quad 
Efficient 3DGS rendering relies on a sorting process that can rapidly re-order each Gaussian primitive based on the update of the camera poses and their clipping planes \cite{06_3DGS}. Therefore, we developed a custom Unity VR renderer through Unity's native render plugin to utilize Kerbl's original CUDA kernels for parallel sorting and tiled-based rendering. As a result, compared to the currently available 3DGS Unity integration \cite{unity3DGSAra}, our renderer can provide performance equivalent to the original CUDA implementation and is better optimized for immersive VR rendering. 

\textbf{Native Renderer Architect}
As our custom renderer does not include the 3D Gaussian points as built-in Unity game assets, we directly displayed the final rendered images as screen quad objects in the HMD. Our 3DGS Unity renderer takes the user's tracked head pose, converts it into the camera view-projection matrix, combines it with other user-defined values such as resolution, FoV, geometric transformation, and updates these parameters in the native  CUDA/C++ renderer at each frame. An effective VR plugin should also correctly synchronize the user's head movement and the rendered images to avoid undesirable temporal aliasing effects such as scene jittering, which occurs in a recent attempt for native CUDA Unity-3DGS integration \cite{CLARTEUnity3DGS}. Our rendering pipeline prevents this problem by triggering native render events inside the \textit{onPreCull} Unity camera event to ensure that all native rendering jobs are completed before displaying the final render texture to users. 

\textbf{Spatial Registration}  To correctly project 3D Gaussians in Unity, we converted the coordinate systems of 3DGS from COLMAP \cite{COLMAP} to Unity. For easy registration of the 3D Gaussians with the Unity world space, we define a reference object in the real world with known scale, rotation, and position whose one-to-one digital copy is available in Unity. Through the reference object, we performed manual calibration to obtain a relative transformation matrix to register the 3DGS model. In addition, we also assume the reference object is the robot's initial position when using the OpenCR's odometry sensor for markless tracking and pose estimation of the robot. 


\begin{figure*}
    \centering
    \includegraphics[width=1\linewidth]{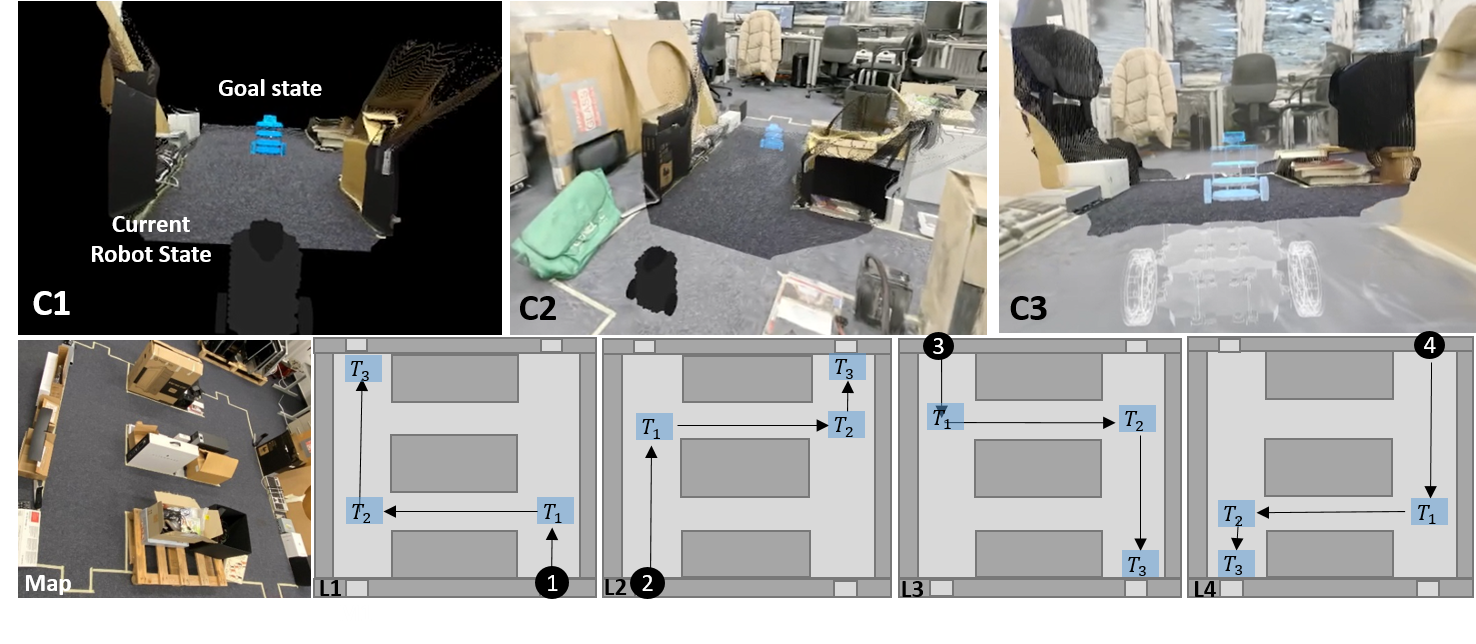}
    \caption{Illustration of our user study experiment design, with C1 showing the exocentric stereo projection condition, C2 showing the exocentric reality fusion condition, C3 showing the egocentric reality fusion condition. The second row illustrates the real-world maze and the four planned teleoperation trajectories.   }
    \label{fig:study_overview}
\end{figure*}

\subsection{Unity Turtlebot Control Module} \label{subsec:turtlebot_control}

\textbf{Overview} \quad We developed a control module in Unity for handling user inputs, managing robot motion, communicating with the remote ROS master, and visualizing real-time robot states. Users can control the robot's movement using the joysticks of the VR controllers with the robot's linear and angular speed linearly mapped to the joystick inputs. The communication module for sending and receiving ROS messages is based on the Unity-TCP-Connector package. Messages with low bandwidth consumption such as odometry and IMU data are sent via the ROS-TCP connector. However, multi-media data such as videos and point clouds are transmitted via UDP for faster processing. To visualize real-time poses of the robot, a digital twin of the robot is rendered whose transform is updated at every frame based on the pose estimation obtained from OpenCR odometry. 

\textbf{Exocentric Control} \quad As Figure \ref{fig:study_overview} C2 demonstrates, our framework allows robust control of the robot from third-person (exocentric) perspective. In the exocentric control mode, an operator can observe the robot's current state in the world from any desired perspective. Users can use the joystick inputs of the VR controllers to translate their positions in the virtual world. While pressing the trigger button of the VR controllers, users can switch the joystick inputs to control the robot's movement instead. 

\textbf{Egocentric Control} \quad 
As Figure \ref{fig:study_overview} C3 demonstrates, our framework also provides an egocentric robot control mode, where the operator sees the environment from the robot's perspective. In egocentric control mode, the user's head position in the virtual environment automatically follows the robot's tracked movement. In addition, we place the user's virtual head position right behind the virtual robot state indicator rather than at the position of the stereo camera so that operator is aware of the robot's position relative to its environment. Notice that in egocentric control mode, we only render half of the digital twin of the robot state indicator, such that the rendering does not obscure the operator's fovea vision of the real-time stereo projection.

\textbf{Other Materials and Software} \quad
Our framework was developed on Unity version 2019.4.29f1 based on OpenGL graphics API and uses the OpenVR desktop runtime and steamVR runtime. In addition, we use the mixed reality toolkit (MRTK) version 2.8 to develop spatial UIs and to manage user inputs in VR. 

\subsection{Overall System Performance}\label{sec:sys_performance}
Our framework can achieve real-time performance on both high-end and moderate workstations. On moderate hardware such as the Alienware m17 R2 laptop with an RTX 2080 graphics card, the overall performance of the entire system, including video decoding, stereo correspondence estimation, stereo projection, point cloud rendering, robot control, and 3DGS rendering runs at $40 - 45 fps$ for an industrial facility environment with $622,335$ Gaussians at $900 \times 960 $ per eye resolution ($50 \%$ resolution of a Meta Quest Pro headset with 106° horizontal FoV, and 96° vertical FoV).  On a high-end device such as the Nvidia RTX 3090 GPU, the overall performance is $30-35$ fps for a room-scale environment with $727,019$ Gaussians at $1536 \times 1440 $ per eye resolution ($80 \% $ of the maximum resolution of a Meat Quest Pro headset with 106° horizontal FoV, and 96° vertical FoV).  

\section{USER STUDY EXPERIMENT}




In this section, we present a user study experiment that benchmarks user performance and evaluates the effectiveness of our framework. The following research questions guided our user study and experiment design: 

\begin{itemize}
    \item \textbf{RQ1}: What is the effect of reality fusion on the operator's cognitive load, task performance, and situation awareness? 
    \item \textbf{RQ2}: For reality fusion, how does the teleoperation perspective change users' performance and perception?
\end{itemize}

\subsection{Conditions}
We used a within-subject experiment design through which participants need to complete a mobile robot navigation task through the following three types of immersive robot teleoperation UIs in VR. Figure \ref{fig:study_overview} C1-C3 presents application screenshots of the three different conditions. 

\begin{enumerate}[label=\textbf{C\arabic*},start=1,itemsep=0mm]
    \item \textbf{Exocentric Stereo Projection Only \cite{01_ego_exo_init_comparison}}: users see only the real-time stereo projection while navigating the robot and their position in the virtual world in exocentric control mode. 
    \item \textbf{Exocentric Reality Fusion}: users see both the real-time stereo projection and 3DGS rendering while navigating the robot and their position in the virtual world in exocentric control mode. 
    \item \textbf{Egocentric Reality Fusion}: users see both the real-time stereo projection and the 3DGS rendering while navigating the robot in egocentric control mode. 
\end{enumerate}
C1 is a reference VR system as proposed by Freland et al. as our comparison baseline \cite{01_ego_exo_init_comparison}. As abundant previous research already revealed the superiority of robot teleoperation systems with VR HMD compared to conventional 2D displays and videos \cite{01_ego_exo_init_comparison, VR-system-mesh}, this study focuses on comparison across different immersive VR robot teleoperation designs only. The ordering of the three conditions for each participant is counter-balanced using a balanced Latin Square method to compensate for carry-over effects.

\subsection{Participants}
We invited 24 participants (10 female and 14 male) to make sure that each condition and route can be equally balanced. Two participants were between 18-24 years old, 17 were between 25-34 years old, 3 were between 35-44, 1 was between 45-54, and 1 was 65 years old or higher.  All were students, researchers, or scientists in HCI, computer science, physics, or robotics. 10 participants use VR systems regularly (at least once a month), and 8 rarely use them ( once or less than once a year). 10 participants never operated a robot before, 11 rarely operated a robot before, and 3 participants worked with a robot regularly. All participants had normal or correct to normal vision. 

\subsection{Tasks}

As illustrated in Figure \ref{fig:study_overview}, we designed a $2.2m \times 2.2m $ maze with four symmetrical different entrance points. Inside the maze, there are three $0.6m \times 0.15 m $ obstacle areas which form two $0.6m \times 0.875m $ pathways. We designed four different trajectories through which users can navigate the robot from one entrance of the maze to the target exit. Each trajectory consists of three subpaths and the level of difficulty for navigation of these subpaths is the same according to the steering law which can predict the amount of time ($T$) users need to navigate through a 2D tunnel given the width of the tunnel $W$ and the length of the tunnel $A$: $T = a+b \frac{A}{W}$ \cite{steeringlaw}. As Figure \ref{fig:study_overview} C1-C3 presents, a goal state indicator (rendered as blue) is presented to inform the participants of the target robot position they need to navigate the robot to. Participants need to sequentially navigate the robot to reach all three goal states ($T1, T2, T3$) in the designated trajectory for each condition. For each different condition, participants are assigned a different navigation trajectory. The maximum speed of the robot was adjusted to $0.05 m/s$ linearly and $0.5 rad/s$ angularly to ensure teleoperation safety.

\subsection{Materials}

The experiment was performed on a Meta Quest Pro headset and a Windows PC with a Nvidia 3090 GPU. The 3DGS model of the real-world maze was reconstructed from 69 images with $3990 \times 2985$ resolution. We generated the 3DGS model following the original model training pipeline developed by Kerbl et al. \cite{06_3DGS}. The training output was post-processed using the Unity 3DGS 2D editing toolkit developed by Pranckevičius \cite{unity3DGSAra}, where we removed outliners and erroneous results to improve the overall visual appearance. The post-processed 3DGS model has $727, 019 $ Gaussians representing around $3.7 m \times 3.7 m \times 1.5 m$ real-world volume which covers the entire maze and its surroundings. The robot teleoperation application runs at $30-35 fps$ at $1536 \times 1440$ per eye resolution as recorded in \ref{sec:sys_performance}.
 
\begin{figure*}
    \centering
    \includegraphics[width=1\linewidth]{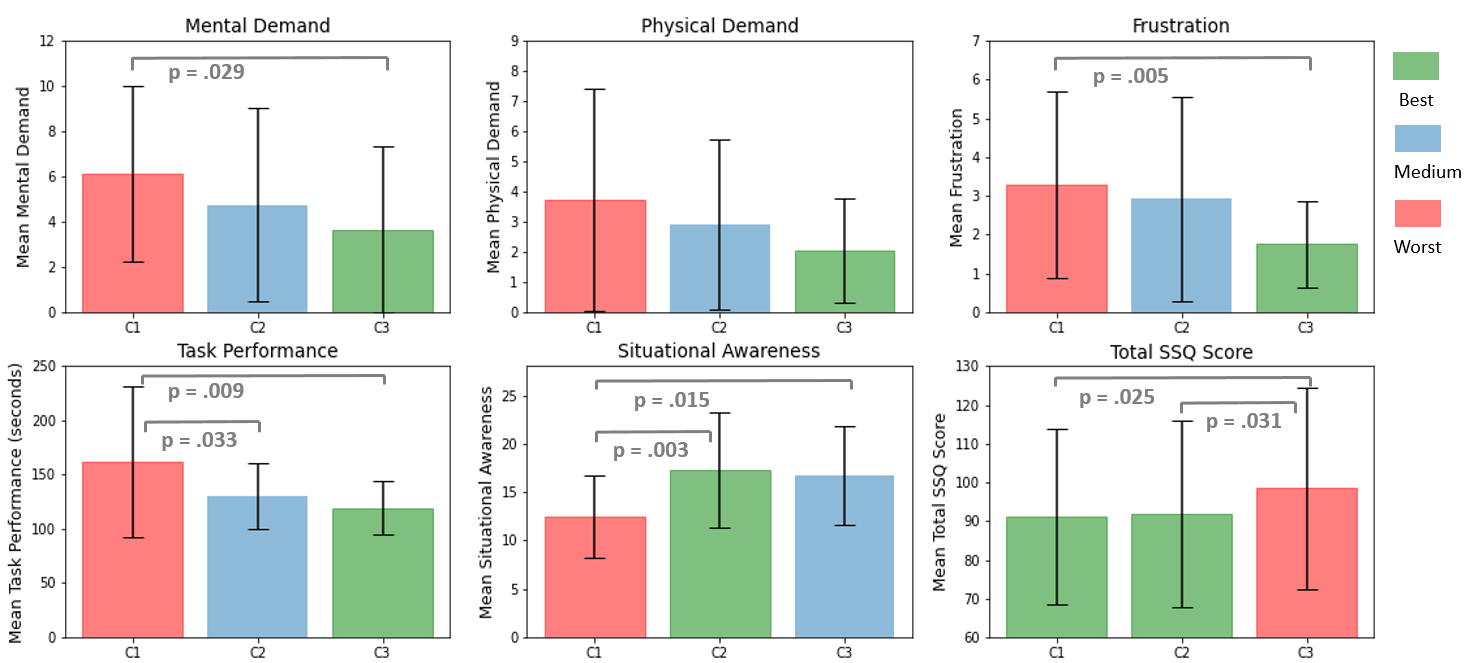}
    \caption{Mean mental demand, physical demand, frustration, task performance, situational awareness, and SSQ score per condition. Vertical bars represent the standard deviation. Any significant differences were labeled with their corresponding p values between conditions.} 
    \label{fig:results_plot}
\end{figure*}
\subsection{Measures} 

\textbf{Task Performance}\quad 
To objectively compare users' performance in different conditions, we record the total elapsed time for users to complete each condition, starting from the moment when the first goal state indicator was displayed in the HMD until the robot successfully reaches the last goal. 

\textbf{Perceived Workload}\quad 
For evaluating users' subjective perceived task loads, we use the standard NASA-TLX questionnaire, which measures various aspects of workload \cite{NASA_TLX}.


\textbf{Situation Awareness}\quad
For evaluating users' situation awareness of the remote environment, we used the Situation Awareness Rating Technique (SART) questionnaire \cite{SART_Taylor_1990}. 

\textbf{Cybersickness}\quad
To measure the amount of induced cybersickness, we use a standard Simulator Sickness Questionnaire (SSQ). The questionnaire was completed before the user study and immediately after each VR exposure \cite{SSQ}. 


\subsection{Hypothesis} \label{sec:hypothesis}
We formulate the following hypothesis concerning previously described measures and conditions: 

\begin{enumerate}[label=\textbf{(H\arabic*)}]
\item Reality Fusion (C2, C3) leads to less perceived cognitive workload while improving the operator's situation awareness and overall performance.

\item Egocentric teleoperation (C3) results in better user performance and lower task load than exocentric teleoperation (C2). However, due to non-self-induced motion, C3 leads to higher motion sickness than C2.

\end{enumerate}

\section{RESULTS and DISCUSSION}
In this section, we present a summary of statistically significant results and discuss their design implications. 

As Figure \ref{fig:results_plot} plots, Wilcoxon signed-rank tests with Bonferroni-Holm adjustment shows that C1 (Point Cloud Only) results in significantly higher mental demand ($Z = -2.584, p = .029, r = -.527$) and a higher level of frustration ($Z = -3.162, p = .005, r = -.645$) than C3 (Egocentric Reality Fusion).  Moreover, Wilcoxon signed-rank tests with Bonferroni-Holm adjustment show that participants perform significantly worse in C1 (Exocentric Reality Fusion) than C2 (Point Cloud Only) ($Z = -2.543, p = .033, r = -.519$) and C3 ( $Z = -2.971, p = .009, r = -.606$). 

In terms of cybersickness, Multiple Wilcoxon signed-rank tests show that C3 (Egocentric Reality Fusion) leads to significantly higher nausea ($Z = -2.539, p = .033, r = -.518$) and overall motion sickness symptoms ($Z = -2.633, p = .025, r = -.537$) than C1 (Point Cloud Only). The overall SSQ score was calculated on a 4-point Likert scale (ranging from 1 - none to 4 - severe). Moreover, C3 (Egocentric Reality Fusion) also leads to significantly higher overall motion sickness than C2 (Exocentric Reality Fusion) ($Z = -2.565, p = .031, r = -.524$). A repeated-measures ANOVA test and Post-hoc tests with adjustment show that participants obtained significantly higher situation awareness in C3 than in C1 ($p=0.015$) as well as higher situation awareness in C2 than in C1 ($p=0.003$). 

In the post-study questionnaire, 12 participants indicated clear preferences for C2, 8 participants indicated clear preferences for C3, 2 indicated equal preferences for C2 and C3, and 2 did not indicate any clear preferences.  

\subsection{Improved Performance with Reality Fusion (H1)} 

As plotted in Figure \ref{fig:results_plot}, statistical analysis confirmed that reality fusion (C2, C3) results in significantly higher situation awareness and better task performance, partially confirming H1. According to participants' qualitative feedback, reality fusion enables ``\textit{a better overview/understanding of the whole environment}" (N=7), makes it easier to ``\textit{determine the robot's global position}" (N=3), and therefore makes it ``\textit{easier to plan routes}" (N=4). Without reality fusion, participants were ``\textit{not sure if there was a direct route between the robot and the target}" and therefore tend to ``\textit{scan around the environment much more}", resulting in worse task performance. 

However, although C2 and C3 lead to lower overall mental demand, physical demand, and frustration, a comparison between C1 and C2 alone does not reveal a significant difference in perceived task load. Therefore, this part of H1 can not be confirmed. This indicates that displaying a global 3D map may introduce extra cognitive demands to users while these extra efforts help them achieve better task performance and gain more situational awareness. 

In terms of SSQ, participants experienced only none to minor motion sickness in both C1 and C2, despite that users were exposed to more complex virtual environments in C2. This is attributed to our technical implementation, which is highly optimized for high-framerate and high-resolution rendering with low video streaming latency. Therefore, it is also safe to conclude that by using reality fusion, users can teleoperate robots in VR in real-time without discomfort. 


\subsection{Exocentric and Egocentric Comparison (H2)}\quad 
In answering RQ2, we compare the results of different measures between C2 and C3. Although C3 leads to lower overall mental demand, physical demand, frustration, and better task performance than C2, the differences were not significant. Therefore, the first part of H2 can not be confirmed. Nonetheless, participants mentioned in the qualitative feedback that it is more demanding to have to control both the robot's movement and their own movements in the virtual environment (N=4). In addition, they found the egocentric teleoperation mode (C3) more immersive and it helps them ``\textit{pay more attention to the real-time point cloud}" and become ``\textit{more aware of the (robot's) environment}". 

The post-study qualitative questionnaire also revealed participants' split preferences for the two teleoperation modes, with those who preferred C3 believing that egocentric teleoperation is more ``\textit{natural}" and ``\textit{intuitive}" (N=2), while others preferring moving freely (N=3) and looking for the best perspective (e.g. a top-down view) on their own. 
 
In terms of SSQ, as expected, participants reported stronger motion sickness symptoms in the egocentric mode due to continuous non-self-induced motion, with a significant difference between C3 and C2 in total SSQ score confirming the second part of H2.  This indicates that while the egocentric teleoperation mode presents certain advantages, it might not be suitable for long-duration teleoperation tasks and could be offered as an option the user could switch to, rather than the main teleoperation mode. 

\section{CONCLUSION and FUTURE WORK}
In this paper, we presented a novel immersive robot teleoperation framework that allows natural, intuitive, and robust remote control of mobile robots in complex semi-structured environments through the reality fusion technique. Our open-source implementation includes a high-performance Unity application for high resolution, photorealistic 3DGS VR rendering, low-latency point cloud streaming, and intuitive mobile robot motion control, as well as a telepresence mobile robot system design that can be easily replicated. We thoroughly evaluated various human aspects of our framework with 24 participants and demonstrated the significant improvement of reality fusion in objective task performance as well as perceived situation awareness.  

In future work, we encourage researchers to enhance reality fusion methods by integrating dynamic SLAM capturing techniques, allowing for real-time updates of the 3DGS global environments to accommodate dynamic changes. A major limitation of the current system is its reliance on the OpenCR board's odometry for tracking the robot's poses, which can lead to motion drift and accumulated tracking errors. Additionally, the current system uses the camera's built-in tracking for projecting point clouds rather than the robot's odometry. This can result in visual misalignment and motion control errors, particularly in environments with uneven floors where the robot's odometry tracking may fail. To improve tracking accuracy, we recommend incorporating unified adaptive tracking and control methods for both the robot and the stereoscopic camera. This can be achieved using vision markers, dynamic SLAM, or optical tracking systems in future implementations. 
\addtolength{\textheight}{-12cm}   






\bibliographystyle{abbrv-doi}

\bibliography{template}

\end{document}